# Near-Optimal Target Learning With Stochastic Binary Signals

**Mithun Chakraborty**
Computer Science Dept.
Rensselaer Polytechnic Institute
Troy, NY 12180, USA
chakrm@cs.rpi.edu

**Sanmay Das**
Computer Science Dept.
Rensselaer Polytechnic Institute
Troy, NY 12180, USA
sanmay@cs.rpi.edu

**Malik Magdon-Ismail**
Computer Science Dept.
Rensselaer Polytechnic Institute
Troy, NY 12180, USA
magdon@cs.rpi.edu

## Abstract

We study learning in a noisy bisection model: specifically, Bayesian algorithms to learn a target value $V$ given access only to noisy realizations of whether $V$ is less than or greater than a threshold $\theta$. At step $t = 0, 1, 2, \ldots$, the learner sets threshold $\theta_t$ and observes a noisy realization of $\text{sign}(V - \theta_t)$. After $T$ steps, the goal is to output an estimate $\hat{V}$ which is within an $\varepsilon$-tolerance of $V$. This problem has been studied, predominantly in environments with a fixed error probability $q < \frac{1}{2}$ for the noisy realization of $\text{sign}(V - \theta_t)$. In practice, it is often the case that $q$ can approach $\frac{1}{2}$, especially as $\theta \to V$, and there is little known when this happens. We give a pseudo-Bayesian algorithm which provably converges to $V$. When the true prior matches our algorithm's Gaussian prior, we show near-optimal expected performance. Our methods extend to the general multiple-threshold setting where the observation noisily indicates which of $k \geq 2$ regions $V$ belongs to.

## 1 INTRODUCTION

Learning with thresholded signals or binary observations is an important problem that appears in many contexts. Typically, a learner who is trying to locate a target identifies a set of intervals, then receives a (potentially noisy) signal indicating which interval the target is in. For example, in online dynamic pricing, a seller wishes to determine the demand curve. She sets a price for a good and observes whether or not the arriving buyer chooses to purchase at that price [Harrison et al., 2010]. In drug dosage discovery, the goal is typically to estimate the maximum dosage level that causes toxicity with less than some target probability (this is typically the focus of Phase I clinical trials) [Cheung and Elkind, 2010]. Threshold queries are also used in image or face localization, where classifiers are used as subroutines to determine whether or not a face or letter or character appears in the query region of some image [Sznitman and Jedynak, 2010].

The literature on learning from such noisy thresholded signals has typically focused on noise of a particular form: nature generates the correct answer, but it is then sent through a noisy transmission channel [Jedynak et al., 2011]. Thus, the probability of seeing the wrong signal is constant, independent of the point of measurement (the particular threshold set by the learner). Several papers have focused on proving the asymptotic optimality of policies that measure either at or around the median [Horstein, 1963, Burnashev and Zigangirov, 1974, Castro and Nowak, 2008]. Recent work shows that measuring at the median is sequentially optimal for entropy reduction in the case of symmetric noise [Waeber et al., 2011]. In a different vein, Karp and Kleinberg [2007] consider noisy binary search: in this problem, a finite sequence of biased coins, ordered in increasing probability of a "heads" outcome, has to be searched for the last element with a probability of heads lower than a specified target value.

The bisection problem itself can also be thought of as a version of the classic problem of stochastic root finding [Robbins and Monro, 1951], where the learner is trying to learn the root of a real-valued, decreasing function $f$. The model is that the learner sequentially queries at points $\theta_1, \theta_2, \ldots, \theta_n$, and receives observations of $f(\theta_1), f(\theta_2), \ldots, f(\theta_n)$ after addition of noise (e.g. zero-mean Gaussian noise). A natural extension to binary signals is to assume that the learner observes whether or not the corrupted signal is above or below zero. This directly corresponds to a noisy binary signal indicating whether the threshold is smaller than or larger than the root. In this case, the noise model is heavily dependent on how close the threshold is set to the target root. When the threshold is near the target, the probability of seeing the wrong signal is significantly higher and no longer bounded away from $\frac{1}{2}$.

We consider learning in exactly this model, assuming a parametric (Gaussian) model for the noise. We discuss the single-threshold case, although our results generalize to multiple thresholds. Specifically, we assume that the learner is trying to locate a target $V$ on the real line. She sets thresholds $\{\theta_t\}_{t\geq 0}$ and receives signals $x_t \in \{-1, +1\}$ generated according to whether or not $V + z_t < \theta_t$ for $z_t \sim_{\text{i.i.d}} \mathcal{N}(0, \sigma_z)$, $t = 0, 1, 2, \ldots$. Little is known about this version of the problem, but lower bounds can be obtained by learning the unthresholded problem, which is just the problem of estimating the mean of a Gaussian by sampling.

**Contributions** We present an algorithm that unconditionally converges to $V$ with high probability. Our algorithm starts with a Gaussian prior on $V$ with mean $\mu_0$ and variance $\sigma_0^2$. Maintaining the posterior is challenging, and we use a moment-matching approach to maintain an approximate posterior which is Gaussian $\mathcal{N}(\mu_t, \sigma_t)$.

The important regime for the algorithm is when the ratio $\rho_0 = \sigma_0/\sigma_z$ is small, because signals are essentially noiseless in the regime where this ratio is high, and any of a variety of algorithms can be used to "localize" the prior appropriately (we give one such algorithm in the paper, and compare it with an exact non-parameteric Bayesian method). If the prior is correct, then we show that, in the $\rho_0 \to 0$ regime, our algorithm outputs an estimate $\hat{V}$ whose expectation (over the prior and the random observations) has converged to within $\varepsilon$ of $V$ in time $O(\frac{1}{\rho_0^2}\frac{1}{\varepsilon^k})$, polynomial in $1/\rho_0$ and $1/\varepsilon$, $k$ being an absolute constant greater than but close to 1, independent of the problem parameters. For comparison, the estimate using exact Bayesian updating with non-thresholded signals has an expectation which has $O(\frac{1}{\rho_0^2}\frac{1}{\varepsilon})$ time to convergence; this shows that our algorithm is asymptotically optimal with respect to $\rho_0$ and near optimal with respect to $\varepsilon$. In other words, noisy binary signals are almost as powerful as the unthresholded original signals.

We also evaluate our algorithm experimentally, and show that its performance in practice is close to that of a learner with access to exact signals. Due to space limitations, we present proof sketches of our main results here and postpone the details to a full version of the paper.

## 2 THE LEARNING PROBLEM

We consider the following sequential learning problem to determine, within error tolerance $\varepsilon$, an unknown value $V \in \mathbb{R}$. At time step $t \in \{0, 1, 2, \ldots\}$, the learner sets a threshold $\theta_t$. In the general setting, $V$ is arbitrary. We model the noisy observations with thresholded additive noise. Specifically, suppose the learner sets a threshold $\theta_t$ at step $t$; the observation $x_t$ is determined by

$$x_t = \text{sign}(V + z_t - \theta_t),$$

where $z_t \sim \mathcal{N}(0, \sigma_z)$ are independent normally distributed noise realizations, and the signum function is $-1$ for a negative argument and $+1$ otherwise. Notice that if $\theta_t = V$, then $x_t$ is a random sign. The learner is faced with two tasks: how to set the threshold at each time step, which is a sequential decision task; and, how to infer $V$ given the observations which result from the thresholds. At some time $T$, the learner outputs an estimate $\hat{V}$ and pays a cost equal to $(\hat{V} - V)^2$.

**The Bayesian Setting.** At time $t = 0, 1, 2, \ldots$, assume a (prior) distribution for $V$, which we denote $p_t(v)$. After observation $x_t$, the Bayesian update to the distribution is given by

$$p_{t+1}(v) = \Phi\left(x_t(v - \theta_t)/\sigma_z\right) p_t(v)/A_t,$$

where $A_t = \int_{-\infty}^{\infty} \mathrm{d}v \;\, p_t(v) \;\, \Phi\left(x_t(v - \mu_t)/\sigma_z\right)$, and $\Phi(\cdot)$ is the standard normal CDF. Assuming $p_t$ is correct, which may not be the case, $p_{t+1}$ incorporates all the new information from $x_t$. At time $t$, the best estimate of $V$ is given by the mean of the distribution, which we define as

$$\mu_t = \mathbb{E}_{p_t}[V] = \int_{-\infty}^{\infty} dv \; v p_t(v).$$

If the learner had to output an estimate for $V$ at time $t$, the expected cost is the variance,

$$\sigma_t^2 = \text{Var}_{p_t}[V] = \int_{-\infty}^{\infty} dv \; v^2 p_t(v) - \mu_t^2.$$

We wish to minimize $\sigma_T^2$. In principle, one can set up a dynamic program to solve this problem, where the state is an entire probability distribution. This is very challenging to solve in such an infinite dimensional space; however, approximate solutions are possible [Das and Magdon-Ismail, 2008]. Further, as we will see, a very simple, myopic strategy works almost as well (provably).

**The Starting Prior.** As with all Bayesian inference algorithms, we need to start with some prior $p_0(v)$. In our context, the noise in the signals is based on a normally distributed random variable $z_t \sim \mathcal{N}(0, \sigma_z)$. One way to quantify the uncertainty in the learner's prior is through the learner's initial variance, which we define as $\sigma_0^2 = \rho_0^2 \sigma_z^2$ (and, in general, $\sigma_t^2 = \rho_t^2 \sigma_z^2$). Given the learner's initial variance, in accordance with the principal of maximum entropy, we adopt the least informative prior. This happens to be the normal distribution, so we assume that $p_0(v) = \mathcal{N}(0, \rho_0 \sigma_z)$ (we can always assume $\mu_0 = 0$ by translating $V$).

The dimensionless parameter $\rho_0$ is an important measure of the harshness of the learning environment. The harshest environment has $\rho_0 \to 0$, where, if the prior is correct, the learner is very sure of her belief about $V$, but the signals are essentially random signs, and so it is hard to make any progress in learning from the observations. This is the regime we are interested in because (i) it is the hard interesting problem; and, (ii) any inference based algorithm will eventually get more and more certain as it receives more observations, which means that $\rho_t \to 0$. Thus, if it is to succeed, any algorithm has to be able to make good progress in this harsh regime. In fact, any reasonable heuristic (and we present one) can learn when the observations are relatively noiseless; the ultimate performance of an algorithm is dependent on its behavior in this $\rho \to 0$ regime. From now on, we set the scale of the problem by choosing $\sigma_z^2 = 1$ (which is without loss of generality).[1]

**Myopic Thresholds.** The simple myopic strategy, within the Bayesian setting, is to set $\theta_t = \mu_t$. This may or may not be sequentially optimal, but as we will prove later, it is sufficient to obtain near-optimal asymptotic performance. With multiple thresholds, the selection of thresholds becomes non-trivial and is dictated by the assumptions/constraints of the problem. However, our state update procedure can easily be extended to such situations and, since multiple thresholds provide strictly more information, the performance cannot be worse than that of the single-threshold algorithm which itself is near-optimal.

To implement the myopic single-threshold strategy, one only needs to compute $\mu_t$ at every time step, and perform the Bayesian update after observing $x_t$. Unfortunately, this computation requires calculation of two integrals (one to compute $A_t$, and one to compute the expectation) which are not analytically tractable, even for elementary starting priors $p_0(v)$. The natural alternative is to use numerical integration. However, numerical integration leads to issues of numerical stability and efficiency. To compute $p_t$, one needs to store the entire history of $\theta_t$, $x_t$, $A_t$, which is $O(t)$, and then the running time to set $\theta_t$, if we compute the integrals numerically with $N$ quadrature points, is $O(Nt)$. Together with the numerical instability, this rapidly becomes computationally infeasible. In addition, algorithmic issues can arise in selecting appropriate finite bounds for the integration domain.

### 2.1 Non-Parametric Histograms

Once the choice of thresholds has been made (in our case, myopically), the main challenge is to efficiently update the prior. A useful near-exact benchmark is to use a non-parametric finite distribution as the prior. Let $v_1 < \cdots < v_N$ be $N$ possible values for $V$ with corresponding probabilities $q_1, \ldots, q_N$. Then the prior $p_0(v)$ is represented by the vector $\mathbf{q}$. The Bayesian update is given by

$$p_{t+1}(v_i) = \frac{1}{A_t} \Phi\left(x_t(v_i - \theta_t)\right) p_t(v_i),$$

where $A_t = \sum_{i=1}^{N} \Phi\left(x_t(v_i - \theta_t)\right) p_t(v_i)$. The mean is computed as a finite sum, $\mu_t = \sum_{i=1}^{N} v_i p_t(v_i)$. Each step takes $O(N)$ time. If we wish to obtain convergence to within $\varepsilon$ of $V$, then the resolution in the finite prior should be $O(\varepsilon)$, which means that $N = \Omega(\varepsilon^{-1})$, making this a computationally intense procedure. Another problem with this approach is that one must commit to a range for $V$, introducing additional assumptions, and leading to serious problems when $V$ is outside, or in the tail of, the range. Nevertheless, this benchmark is useful for giving some insight into the behavior of the Bayesian update, which will lead to our proposed algorithm. Starting from a Gaussian prior, and running the non-parametric updates, we illustrate how the posterior evolves in Figure 1.

### 2.2 Our Algorithm: Approximate Inference

The detailed description of our algorithm is given in Section 3. We give a quick overview of the intuition

[1] The scale can always be added back through powers of $\sigma_z$ using dimensional arguments.

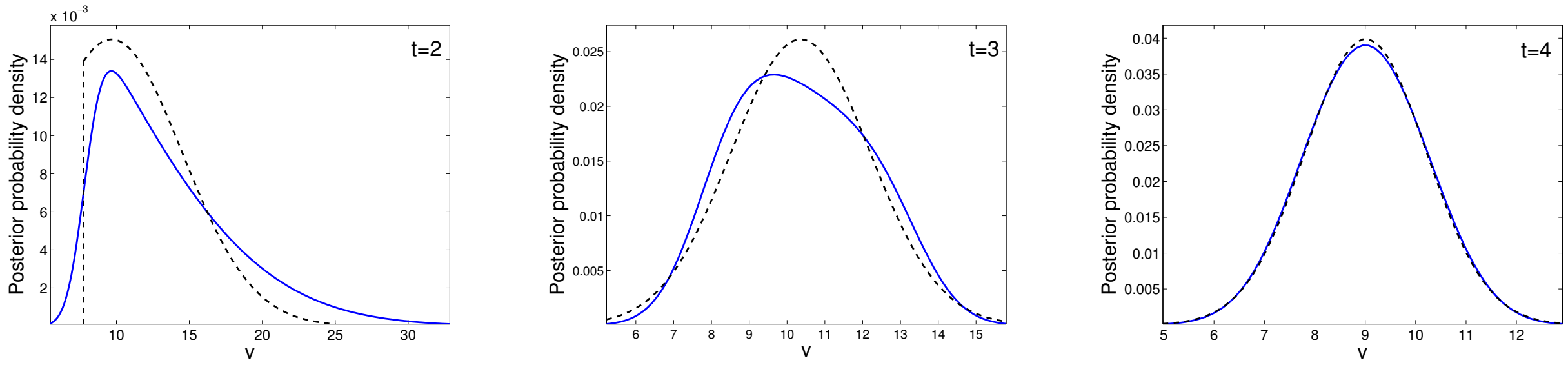

(i) Truncation (ii) Collapse (iii) Convergence

Figure 1: Evolution of $p_t(v)$ using non-parametric histograms and Bayesian updates; $p_0(v)$ is $\mathcal{N}(0, \rho_0)$, $\rho_0 = 10$, $V \approx 9.45$. Typical evolution consists of 3 main phases. (i) **Truncation** ($\rho_0 \gg 1$): all observations are in the same direction (here $x_t = +1$) and almost noiseless, so the Bayesian update essentially truncates and renormalizes the distribution. Shown for comparison is the truncation heuristic from our algorithm, which approximates this phase. (ii) **Collapse**: when the first observation in the opposite direction arrives (here $x_t = -1$), the distribution collapses to something more symmetric, although not quite normal. Shown for comparison is the entropy-matched normal with the same mean. (iii) **Convergence**: $\rho_t$ is typically small and $\mu_t$ is close to $V$ (the algorithm has (probabilistically) bracketed $V$). From then on, nearly independent observations which are close to random signs cause the distribution to rapidly converge to normal (as would be expected with truly independent observations).

here. The basic idea is to start from a normal prior, $p_0(t) = \mathcal{N}(0, \rho_0)$ and perform *approximate* inference, by maintaining approximations to $p_t$ and placing thresholds according to these approximations. We defer proofs for the moment, but briefly mention the nature of these approximations (see Figure 1 for an illustration). First, when all the observations are in the same direction, we maintain a truncated normal distribution. Upon collapse (the first observation in the opposite direction), we revert back to Gaussian, using entropy matching to set its parameters. Though the Gaussian is not very accurate at collapse, this is only a transient, and so is not very important. As convergence occurs, the Gaussian becomes a better and better approximation, so we remain in the Gaussian world, using moment matching to update the parameters.

### 2.3 A Lower Bound for Optimal

We briefly comment on the best possible algorithm within the Bayesian setting under the assumption that the prior $p_0(v) \sim \mathcal{N}(0, \rho_0)$ is correct. Suppose one could actually observe the signals $\hat{x}_t = V + z_t$ (without thresholding). Then one is certainly getting more information and so should be able to do better. With non-thresholded observations, we *can* do the exact inference, because this is just a simplified scalar Kalman filter. Here, the true posterior is Gaussian with

$$\mu_{t+1} = \frac{\mu_t + \rho_t^2 (V + z_t)}{1 + \rho_t^2} \quad \text{and} \quad \rho_{t+1}^2 = \frac{\rho_t^2}{1 + \rho_t^2}. \quad (1)$$

Note that the evolution of $\rho_t$ is deterministic and can be solved for exactly: $\rho_t^2 = \rho_0^2 / (1 + \rho_0^2 t)$. This gives our expected cost at time $T$. One can actually solve for the full distribution of $\mu_t$. The following theorem, which follows using standard probabilistic arguments, gives the convergence of $\mu_t$ both in expectation and with high probability. We assume $\mu_0 = 0$, $\{z_t\}_{t \geq 0}$ are i.i.d. $\mathcal{N}(0,1)$, *i.e.* $\sigma_z = 1$.

**Lemma 1.** *$\mu_t$ has a normal distribution with*

$$\mathbb{E}[\mu_t] = V - \frac{V}{1 + t\rho_0^2} \quad \textit{and} \quad \mathrm{Var}[\mu_t] = \rho_0^2 \frac{t\rho_0^2}{(1 + t\rho_0^2)^2}.$$

Fix an error tolerance $\varepsilon > 0$. The dependence of the expected value on $t$ immediately implies a lower bound on the time after which the expectation of $\mu_t$ (which is our output estimate $\hat{V}$) is within $\varepsilon$ of $V$. Further, the distribution for $\mu_t$ tells us that if we fix a small confidence parameter $\delta$, $0 < \delta \ll 1$, and define $\zeta = -\Phi^{-1}(\delta) = \Theta(\sqrt{\ln(1/\delta)})$, then (for $V > 0$) with probability at least $\delta$, $\mu_t \leq \mathbb{E}[\mu_t] - \zeta\sqrt{\mathrm{Var}[\mu_t]}$, which allows us to get a lower bound on $t$ if we want high probability convergence.

**Theorem 1.** *Fix $\varepsilon < \frac{|V|}{2}$, $\delta \leq \Phi(-1)$. For $t > |V|/2\varepsilon\rho_0^2$, $|V| - |\mathbb{E}[\mu_t]| < \varepsilon$. For $t > \max\left\{\frac{2|V|}{\varepsilon\rho_0^2}, \frac{4\zeta^2}{\varepsilon^2}\right\}$ then $\Pr[\mu_t > V - \varepsilon] > 1 - \delta$ for $V > 0$ and $\Pr[\mu_t < V + \varepsilon] > 1 - \delta$ for $V < 0$.*

The theorem says that to get convergence in expectation, $O(V/\varepsilon\rho_0^2)$ time is needed; if one wants convergence with probability at least $1 - \delta$, then

**Algorithm 1** The Learning Algorithm

Initialize $l_0 = -\infty, r_0 = \infty, m_0 = \mu_0, s_0 = \sigma_0$.
**for** $t = 0, 1, 2, \ldots$ **do**
  Set threshold at $\mu_t$;
  Receive noisy thresholded signal $x_t$;
  Update $l_t, r_t, m_t, s_t$;
  Compute $\mu_{t+1}, \rho_{t+1} = \sigma_{t+1}/\sigma_z$;
**end for**

$O(V/\varepsilon\rho_0^2 + \ln\frac{1}{\delta}/\varepsilon^2)$ time is needed. These bounds will be useful in showing that our algorithm is nearly as good as exact inference with non-thresholded observations. This will mean that thresholding does not significantly impede our ability to learn!

## 3 THE ALGORITHM

Our algorithm is illustrated in Figure 2. The state of the learner at time *t* is completely described in terms of four parameters, $(l_t, r_t, m_t, s_t)$, that describe its current belief distribution, which can take on two forms: either Gaussian, or truncated Gaussian. The support of the distribution is given by $(l_t, r_t)$ (in all cases, either $l_t = -\infty$ or $r_t = \infty$); the location and shape of the distribution are determined by $m_t, s_t^2$ (mean and variance of the underlying Gaussian). Thus the learner's belief distribution is a rescaled normal distribution on the support $(l_t, r_t)$.

$$p_t(v) = \begin{cases} \dfrac{N\left(\frac{v-m_t}{s_t}\right)}{s_t\left(\Phi\left(\frac{r_t-m_t}{s_t}\right) - \Phi\left(\frac{l_t-m_t}{s_t}\right)\right)} & v \in (l_t, r_t), \\ 0 & \text{otherwise,} \end{cases}$$

where $N(\cdot)$ is the standard normal PDF. The initial prior is normal with mean 0 and variance $\rho_0^2$ ($\sigma_z = 1$), which is described by the state $(-\infty, \infty, 0, \rho_0)$. One parameter in the algorithm is a threshold $\rho^*$; as long as $\rho_t \le \rho^*$, the learner's belief will be Gaussian. If the belief is Gaussian and $\rho_t > \rho^*$, the belief will transition to a truncated Gaussian. The side of the truncation depends on the sign of the next observation. If the belief is a truncated Gaussian, then it will transition to Gaussian if the standard deviation $\rho_t$ drops below $\rho^*$; or, if the signal causes the distribution to collapse as indicated in Figure 1 (this happens when either $l_t = -\infty$ and $x_t = +1$ or $r_t = \infty$ and $x_t = -1$). A high level desciption is given in Algorithm 1. Now we describe the detailed update procedures.

**Approximate Gaussian Inference ($\rho_t \le \rho^*$).** As in Das and Magdon-Ismail [2008], in transition-

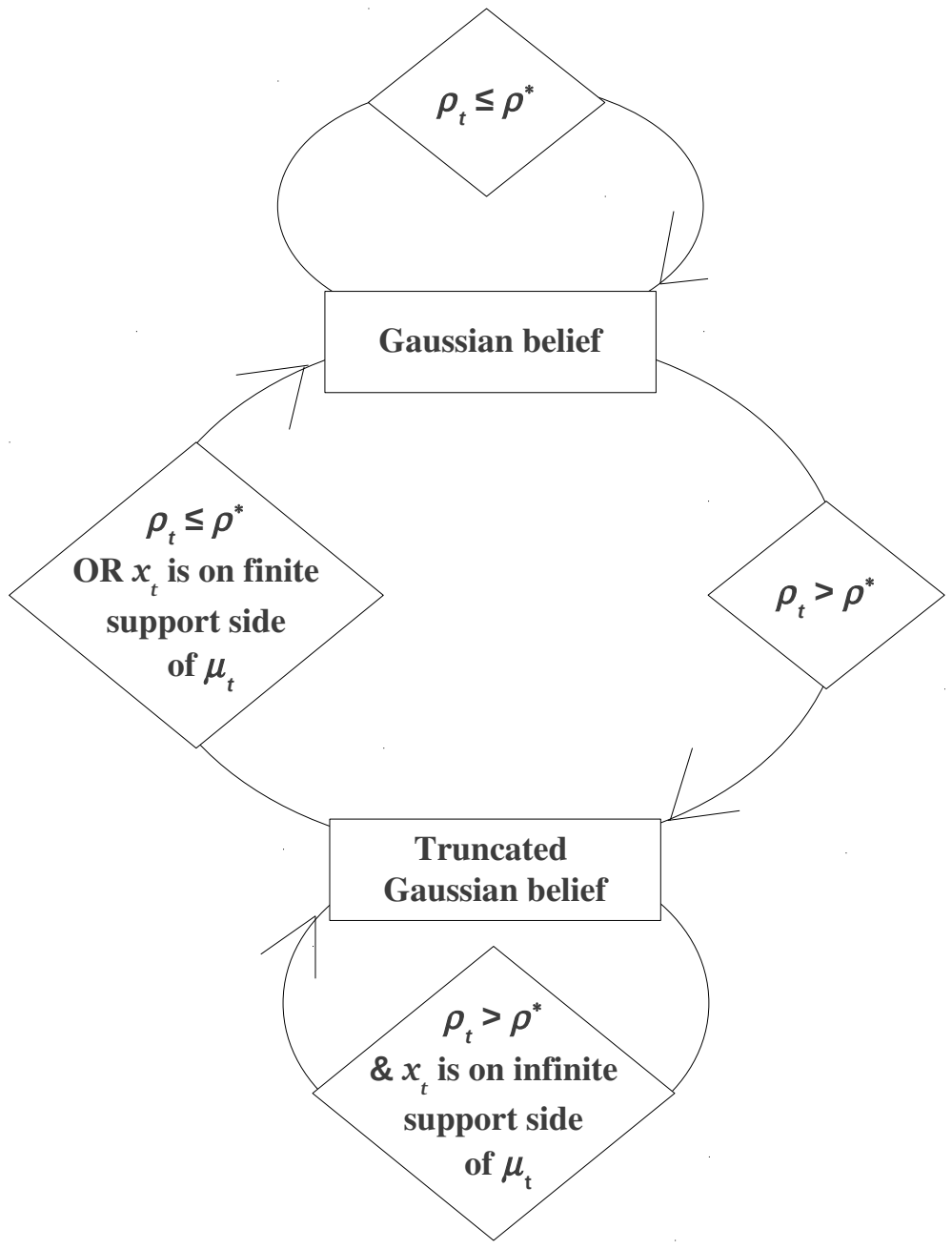

Figure 2: Learner's state transitions.

ing from Gaussian to Gaussian, we can compute the mean and variance of the true posterior, and we approximate this with the Gaussian that has the same mean and variance. So we perform *approximate moment matching inference* in this case. Das and Magdon-Ismail [2008] derive exactly such moment matching equations for two thresholds, which we can directly specialize to the single threshold case:

$$\mu_{t+1} = \mu_t + x_t \frac{\left(\sigma_z\sqrt{2/\pi}\right)\rho_t^2}{\sqrt{1+\rho_t^2}}; \quad (2)$$

$$\rho_{t+1}^2 = \rho_t^2\left[\frac{1+\rho_t^2(1-2/\pi)}{1+\rho_t^2}\right]. \quad (3)$$

**Truncation ($\rho_t > \rho^*$).** When $\rho_t$ is large, we approximate the inference by truncating (as in Figure 1) as long as the signal is consistent with the truncation. The state updates are:

$$(l_t, \infty, m_t, s_t) \overset{(\theta_t, x_t = +1)}{\longrightarrow} (\theta_t - 2\sigma_z, \infty, m_t, s_t);$$
$$(-\infty, r_t, m_t, s_t) \overset{(\theta_t, x_t = -1)}{\longrightarrow} (-\infty, \theta_t + 2\sigma_z, m_t, s_t).$$

**Collapse** No matter what $\rho_t$ is, if the signal is inconsistent with the truncated Gaussian, then we collapse back to Gaussian (see Figure 1). Unfortunately, updating to a Gaussian using moment matching would take the distribution with finite support and collapse to a distribution with infinite support and the same variance, typically producing a Gaussian that is too localized although there can be a lot of uncertainty in the learner's posterior. So a better way to capture this uncertainty is by matching the entropy. We call this approximate inference by entropy matching. To make the entropy matching analytically tractable, we first doubly truncate the distribution (as in regular truncation), compute the mean and entropy of the resulting distribution, and then collapse to the Gaussian with this mean and entropy. For finite $l_t$ and $r_t$, the state updates are:

$$(l_t, \infty, m_t, s_t) \overset{(\theta_t, x_t=-1)}{\longrightarrow} (-\infty, \infty, \mu_{t+1}, \rho_{t+1});$$
$$(-\infty, r_t, m_t, s_t) \overset{(\theta_t, x_t=+1)}{\longrightarrow} (-\infty, \infty, \mu_{t+1}, \rho_{t+1}).$$

We abuse notation above, in that the updates are not the same in both cases. In the top case ($x_t = -1$), set $l = l_t$ and $r = \mu_t + 2\sigma_z$; in the bottom case ($x_t = +1$), set $l = \mu_t - 2\sigma_z$ and $r = r_t$. Then, a tedious but straightforward computation of the mean and entropy of the resulting rescaled doubly truncated Gaussian with support $(l, r)$ and parameters $(m_t, s_t)$, followed by entropy matching gives:

$$\mu_{t+1} = m_{t+1} + s_{t+1}\left[\frac{N(l') - N(r')}{\Phi(r') - \Phi(l')}\right],$$
$$\sigma^2_{t+1} = s^2_{t+1}\left[(\Phi(r') - \Phi(l'))^2 e^{\frac{l'N(l') - r'N(r')}{\Phi(r') - \Phi(l')}}\right].$$

where $l' = (l - m_t)/s_t$, $r' = (r - m_t)/s_t$.

# 4 ANALYSIS

We now analyze the algorithm presented in Section 3. The algorithm has two basic phases. The first is if $\rho_t$ is large, in which case the algorithm is a heuristic that truncates the distribution until colapse into the Gaussian world, at which point the process (truncation→collapse) repeats until $\rho_t$ gets below $\rho^*$. We do not go into the details of the dynamics for $\rho_t > \rho^*$ because in this relatively noiseless regime, many heuristics can "localize" the posterior quickly. The interesting regime is $\rho_t \to 0$, when the signals start to get noisy. In this regime, our algorithm will always be doing approximate Gaussian inference (since $\rho_t$ is decreasing), updating according to Equations (2) and (3). Once in this regime, we essentially show that our algorithm is near optimal by proving that $\mu_t$ converges quickly to $V$ in expectation, and it also does so with high probability. For asympotic results, we have in mind that $\rho_0 \to 0$.

**Theorem 2.** *There exist absolute positive constants $C > 0$ and $k$, $1 \le k < \pi\sqrt{2} \approx 4.443$ such that, if $t > C/(\rho_0^2 \varepsilon^k)$, then $|V| - |\mathbb{E}[\mu_t]| < \varepsilon$.*

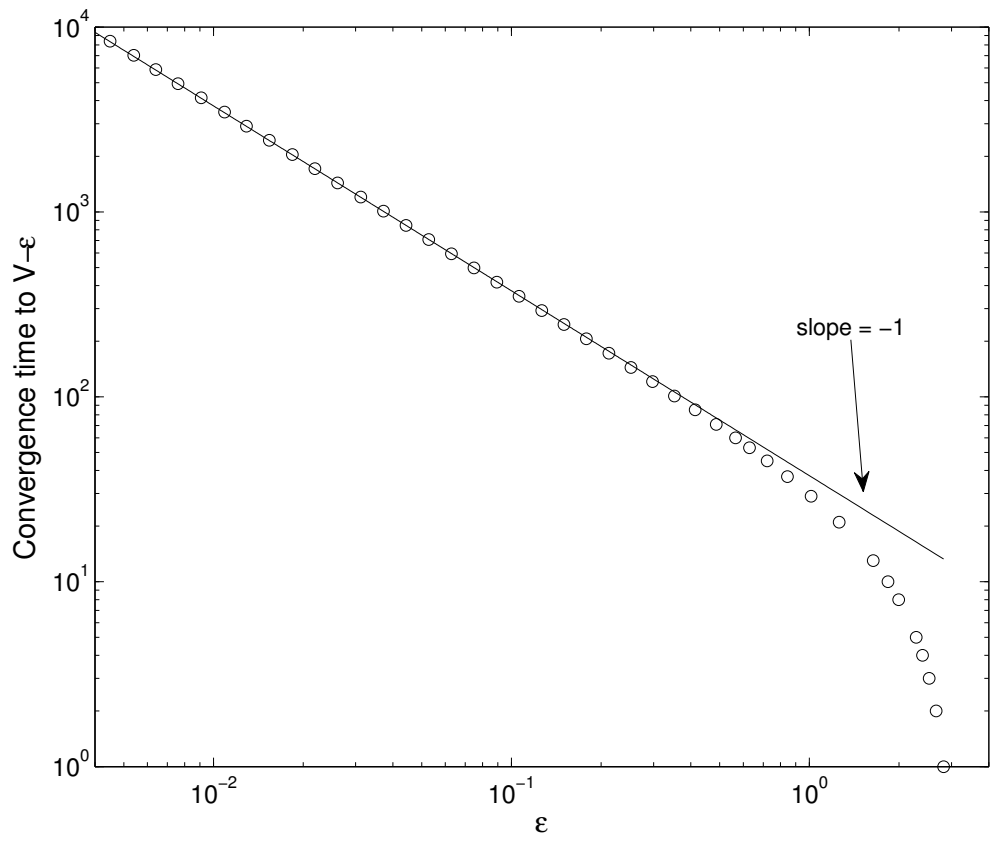

Figure 3: An example of the time for $\mathbb{E}[\mu_t]$ (computed over $10^6$ simulations) to converge to $V - \varepsilon$ (log-log scale) when $V = 3$, $\mu_0 = 0, \rho_0 = 0.5$. Evidently, the convergence time approaches $O(1/\varepsilon)$ as $\varepsilon$ becomes small, hence it is near optimal.

Recall that the expectation is with respect to $p_0(v)$ and the i.i.d. $z_t \sim \mathcal{N}(0, 1)$, hence this result is relevant when the prior is correct. From Theorem 1, the best we could hope for, even with non-thresholded signals, for the expectation to get within $\varepsilon$ of $V$ is $t = O(1/\rho_0^2\varepsilon)$. Thus, our dependence on $\rho_0^2$ is optimal. The theorem gives polynomial convergence in $1/\varepsilon$ but, in practice, $k$ is almost 1, which is near optimal asymptotic convergence, as illustrated with an example in Figure 3.

Our second result demonstrates unconditional convergence with high probability, regardless of whether our prior $p_0(v)$ is correct. For simplicity, we assume without loss of generality that $V > 0$. Note that $\mu_t$ follows a stochastic process. We ask how long we have to wait before, with high probability, $\mu_t$ will have crossed $V - \varepsilon$. This analysis is sufficient to convey the main point of the convergence, because once you cross this barrier, the stochastic process has an attractor at $V$, and so will stay in this region. The tough part is getting to this region.

**Theorem 3.** *Fix $0 < \delta < 1$, $0 < \varepsilon < V$, $0 < \rho_0 \le 1$, and define $\Delta = V - \varepsilon$. There is an absolute constant $C > 0$ such that if $t > T = e^{C(\ln(1/\delta) + \Delta)/\varepsilon}/\rho_0^2$, then with probability at least $1 - \delta$, $\max_{i \le t} \mu_i > V - \varepsilon$.*

This constant $T$ gives us an upper bound on the time at which $\mu_t$ first crosses $V - \varepsilon$. In the practical setting where the prior is extremely ill specified, $V$ is very large and $\varepsilon$ is usually specified as a percentage of $V$. Then, the exponent is some constant and the dependence on $1/\rho_0$ is what we are interested in. Comparing with Theorem 1, we see that our algorithm is asymptotically optimal with respect to $1/\rho_0$.

We now sketch the proofs of Theorems 2 and 3. Throughout, assume that $\rho_0 \leq 1$. Let $\eta_t$ be the magnitude of the change in $\mu_t$ at time $t$. The proofs of both theorems rely on the following lemma which shows that $\eta_t = \Theta(1/t)$.

**Lemma 2.** *For $\rho_0 \leq 1$, $\eta_t < \frac{c_2}{t}$ and*

$$\eta_t > \begin{cases} c_1\rho_0^2 & t \leq \lfloor 1/\rho_0^2 \rfloor, \\ \frac{c_1}{t} & t \geq \lfloor 1/\rho_0^2 \rfloor + 1, \end{cases}$$

*where $c_1 = \frac{1}{2\sqrt{\pi}}$, $c_2 = \sqrt{\frac{2}{\pi}} \left( \frac{\pi + (\pi-2)\rho_0^2}{2} \right)$.*

**Proof of Theorem 2 (sketch).** Let $V > 0$. Define $\xi_t = \mathbb{E}[\mu_t]$ as the expectation of $\mu_t$ with respect to $p_0(v)$ and $\{z_i\}_{i<t}$ and $\mathbb{E}_{x_t}[\cdot]$ as the expectation with respect to the input $x_t$ at time $t$. Now, $\mathbb{E}_{x_t}[\Delta\mu_t] = (2p_t^+ - 1)\eta_t$, where $p_t^+ = \Pr[x_t = +1] = \Phi(V - \mu_t)$. The basic idea is to observe that $2\Phi(V-\mu_t) - 1 = O(V - \mu_t)$, and so $\xi_{t+1} = \xi_t + O(V - \xi_t)\eta_t$, or $\xi_{t+1} - \xi_t \approx c(V - \xi_t)/t$ for some constant $c$, since $\eta_t = \Theta(1/t)$. If we treat this difference equation approximately as a differential equation and integrate, we find that the time needed for $\xi_t$ to reach $V - \varepsilon$ satisfies $t = t^* \left(\alpha/\varepsilon\right)^k$, $k = 1/c$, where $\alpha > 0$ is a constant independent of $\varepsilon$, $t^*$ is the time needed for all the approximations to become accurate and is asymptotically $O(e^{V/\alpha}/\rho_0^2)$. However, asymptotically, $\eta_t$ gets close to its upper bound of $\sqrt{\frac{2}{\pi}} \left( \frac{\pi + (\pi-2)\rho_0^2}{2} \right)/t$ so that $k$ approaches its lower bound of 1.

**Proof of Theorem 3 (sketch)** Given the initial belief distribution $\mathcal{N}(0, \rho_0)$, the value of $\rho_t$, and hence of $\eta_t$, for each $t \geq 0$ is completely determined. Thus, after $t$ time-steps, the learner could attain any one of at most $2^t$ pre-defined values of $\mu_t$ each corresponding to a unique path of the form $\left[(0,0), (\mu_{(1)}, 1), ..., (\mu_{(t)}, t)\right]$ in the $(\mu, t)$-space, where $\mu_{(t)}$ denotes one of the possible mean beliefs that the learner could have at time $t$. With this insight, we define a reinforcement learning setting in which each such path is a state of the learner. Define $\mathcal{S} = \{s = \left[(0,0), ..., (\mu_{(t)}, t)\right]; \mu_{(t)} < V - \varepsilon\}$. Obviously, $p_t^+ > \hat{p} = \Phi(\varepsilon) > 0.5$ for any $s$ in $\mathcal{S}$.

Let us denote by $\pi$ the policy under which the action in any state $s \in \mathcal{S}$ is to go to the state $[s; (\mu_{(t)} + \eta_t, \rho_{t+1})]$ or $[s; (\mu_{(t)} - \eta_t, \rho_{t+1})]$ with constant probabilities $\hat{p}$, and $(1-\hat{p})$ respectively, and by $\pi'$ the policy under which the corresponding transition probabilities are $p_{(t)}^+$ and $(1 - p_{(t)}^+)$ ($\pi'$ corresponds to our approximate inference algorithm). For a given time-horizon $[0, \tau]$, the value function $V_\tau^{\pi^*}(s)$ for state $s$ under policy $\pi^* \in \{\pi, \pi'\}$ is defined as the probability of hitting $(V - \varepsilon)$ in the interval $[t, \tau]$, starting from state $s$ under the policy $\pi^*$.

**Lemma 3.** *For any $s = \left[(0,0), ..., (\mu_{(t)}, t)\right] \in \mathcal{S}$, where $t < \tau$, $V_\tau^{\pi'}(s) \geq V_\tau^{\pi}(s)$.*

In particular, for the initial state $\varphi = [(0,0)]$, $V_\tau^{\pi'}(\varphi) \geq V_\tau^{\pi}(\varphi)$. We now focus on deducing a lower bound on $V_\tau^{\pi}(\varphi) = \Pr\left[\mu_{(\tau)} \geq V - \varepsilon | \pi\right]$ for the dominating process. Noting that for any $s \in \mathcal{S}$ under $\pi$, $\mathbb{E}[\Delta\mu_t] = (2\hat{p} - 1)\eta_t \ \forall t \geq 0$ so that $\mu_{(t)}$ is the sum of i.i.d. random variables $\{\Delta\mu_i\}_{i=0}^{t-1}$ and has expectation $(2\hat{p} - 1)\sum_{i=0}^{t-1}\eta_i$, $\Pr\left[\mu_{(\tau)} \geq V - \varepsilon | \pi\right]$ equals

$$\Pr\left[\mu_{(\tau)} - \mathbb{E}\left[\mu_{(\tau)}\right] \geq -(2\hat{p} - 1)\sum_{t=0}^{\tau-1}\eta_t + \Delta \Big| \pi\right]$$

where $\Delta = V - \varepsilon$. It is not difficult to show that for $\tau > \tau' = \left(\lfloor \frac{1}{\rho_0^2} \rfloor + 1\right) \exp\left(\frac{2\sqrt{\pi}\Delta}{2\hat{p}-1}\right)$, $(2\hat{p}-1)\sum_{t=0}^{\tau-1}\eta_t - \Delta > 0$ . Hence, for $\tau > \tau'$, by Hoeffding inequality, we have that $\Pr\left[\mu_{(\tau)} \geq V - \varepsilon | \pi\right] \geq 1 - \delta$ for

$$\delta \geq \exp\left[-\frac{2\left((2\hat{p}-1)\sum_{t=0}^{\tau-1}\eta_t - \Delta\right)^2}{\sum_{t=0}^{\tau-1}(2\eta_t)^2}\right].$$

Combining this result with Lemma 3, we conclude that the above inequality also holds for our algorithm, *i.e.* $\Pr\left[\mu_{(\tau)} \geq V - \varepsilon | \pi'\right] \geq 1 - \delta$. After some tedious algebra, we can finally obtain the required lower bound on the time for the above inequality to hold for a given $\delta$.

# 5 EXPERIMENTAL RESULTS

We perform experiments to evaluate the practical performance of our algorithm. Our simulations measure convergence time as the time taken by $\mu_t$ to enter the region $[V - \varepsilon, V + \varepsilon]$ for the first time. We are interested in the dependence of the convergence time on $\rho_0$ and $\varepsilon$.

First, we compare the non-parametric algorithm (**NonParam**) to exact inference on non-thresholded signals, and show that noisy binary signals are almost as informative as the unthresholded signals. This is already surprising. Assuming that the prior is correct, we set the support of the non-parametric histogram to $[-6\rho_0, 6\rho_0]$ and use 1,000 histogram bins. We generate $V$ randomly according to $p_0$ and any value of $V$ outside this finite support is discarded. For our algorithm, we set the switch-over parameter $\rho^* = 2.5$. The number of steps taken by each algorithm to converge to the region $[V-\varepsilon, V+\varepsilon]$, averaged over $10^8$ runs for **NonParam** and $10^9$ runs for each other algorithm, is reported in Figure 4.

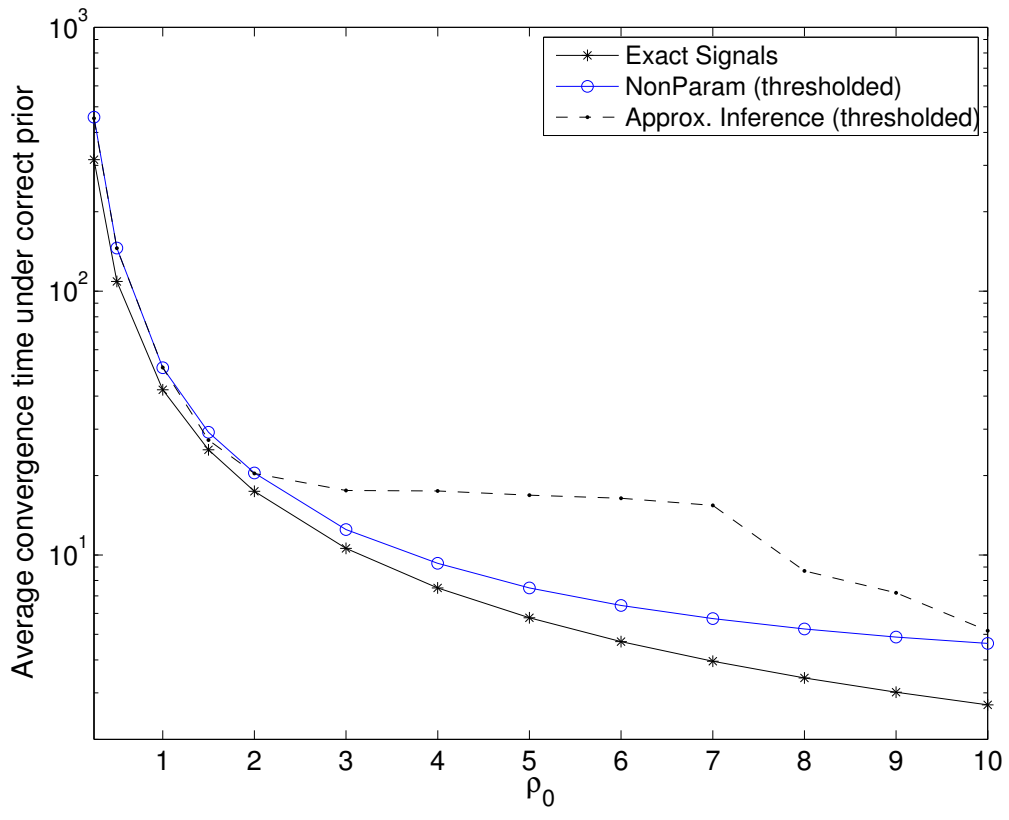

Figure 4: Plot of average correct-prior convergence time *vs* $\rho_0$, logarithmic along the vertical axis.

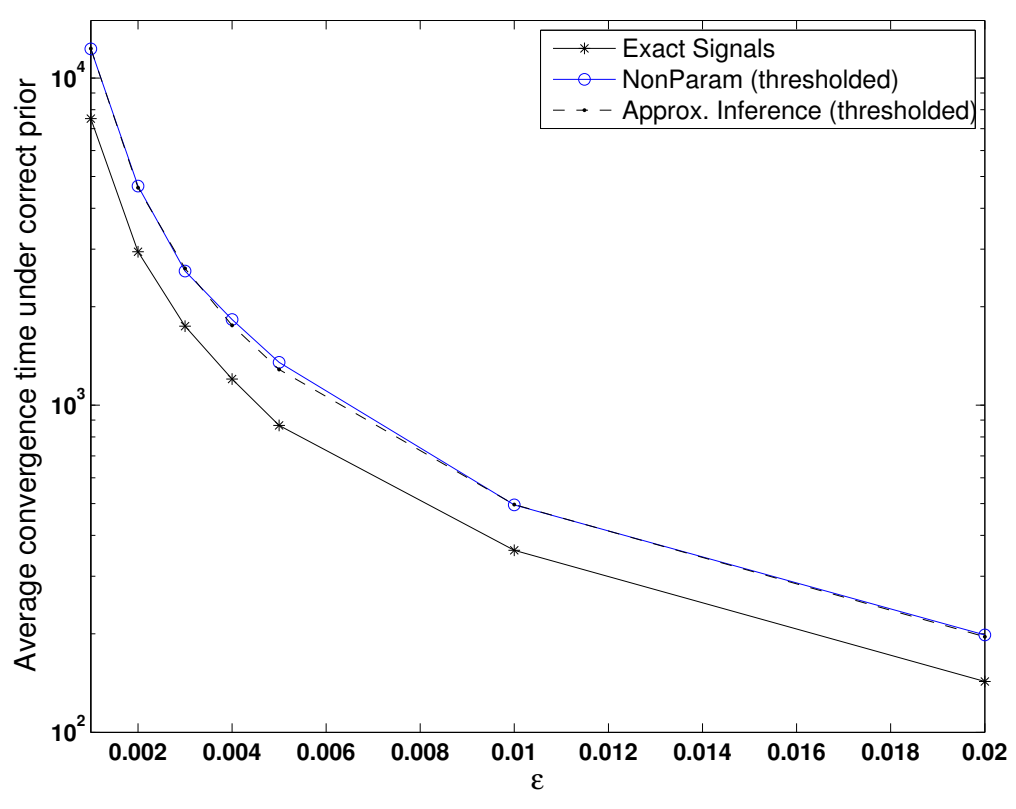

Figure 5: Plot of average correct-prior convergence time *vs* $\varepsilon$, logarithmic along the vertical axis.

In our second set of simulations, we fix $\sigma_0$ at 0.5 and vary $\varepsilon$. To ensure adequate resolution for **NonParam**, we use $24\rho_0/\varepsilon$ bins (giving a resolution of $\varepsilon/2$). The number of steps to convergence is presented in Figure 5. The average is over $10^5$ runs for **NonParam** (owing to computational burden) and $10^7$ runs for each other algorithm. It is clear from the figures that not only is learning from noisy binary thresholds feasible in this Bayesian model, but can be performed almost as well as learning from non-thresholded signals, in accordance with the theory.

### Acknowledgements

We thank Peter Frazier for helpful comments. This research is supported by an NSF CAREER Award (IIS-0952918) to Das. Magdon-Ismail is partially supported by the U.S. DHS through ONR grant number N00014-07-1-0150 to Rutgers University. This work does not necessarily reflect any position or policy of the US Government.